# Enhancing the prediction of disease outcomes using electronic health records and pretrained deep learning models

Zhichao Yang,[1] Weisong Liu,[2] Dan Berlowitz,[3,4] and Hong Yu[1,2,4,*]

[1]College of Information and Computer Science, University of Massachusetts Amherst, MA, USA
[2]Center for Biomedical and Health Research in Data Sciences, Department of Computer Science, University of Massachusetts Lowell, MA, USA
[3]College of Health Sciences, University of Massachusetts Lowell, MA, USA
[4]Center for Healthcare Organization and Implementation Research, Bedford VA Medical Center, MA, USA
[*]Corresponding Author: Hong_Yu@uml.edu

## Introduction

Longitudinal electronic health records (EHRs) can be used for enhancing the prediction of individual patient disease or outcome onsets. Early prediction approaches include regression, traditional machine-learning-based, and ensemble models (e.g., support vectors machines, random forest, and gradient boosting).[1] Deep-learning-based models have now been shown to be superior to these earlier models.[2,3]

Most work in predictive models focuses on prediction of single disease or outcome onsets (e.g., sepsis and diabetes). Details of such research are described in the supplementary materials. Such models typically need domain experts for cohort and feature selection, and in most cases will not take advantage of full longitudinal EHRs. However, patients typically do not have a single disease but have multiple comorbidities, many of which are highly correlated, such as obesity, diabetes, and hypertension, and that may collectively contribute to a variety of important outcomes overtime. We hypothesize that a model that is trained to predict patients' multiple comorbidities would optimize its potential for single disease or outcome onset prediction.

In this study, we propose DeCode, an innovative predictive model that is pretrained on a large dataset of longitudinal EHRs to predict the complete diagnostic ICD codes of patients' future visits. We then fine-tuned DeCode for specific clinical disease or outcome prediction. DeCode adapts the Bidirectional and Auto-Regressive Transformers (BART)[4] architecture and differs from BART in innovative noising schemes and medical training objectives, which are the contributions of this study. We evaluated DeCode extensively on predictions of a broad range of clinical disease and outcome onsets, including: pancreatic cancer, intentional self-harm of patients with post-traumatic stress disorder (PTSD), and hospital admission for type 2 diabetes or hypertension. Through extensive evaluations, our results show that pretraining DeCode improved performance on the downstream single-disease or outcome onset prediction tasks.

## Methods

### Data

To train/evaluate DeCode and compare it with the baseline models, we first built a large EHR cohort for pretraining, disease-agnostic onset prediction (DAOP), and single disease/outcome onset prediction. Using the Veterans Health Administration's Clinical Data Warehouse (VA-CDW), we first identified a total of 8,308,742 patients who received care from the Veterans Health Administration from 1/1/2016 to 12/31/2019. After excluding patients with less than 2 visits, the remaining 6,829,064 patients were randomly split 95% (6,475,218) and 5% (353,846) as, respectively, the pretraining dataset and the dataset for the downstream DAOP and single disease/outcome onset predictions. The detailed patient cohort selections and the downstream disease or outcome onsets for this study is shown in Supplementary Figure 1. Supplementary Table 1 shows the detailed statistics for the pretraining dataset.

### Predictors

DeCode takes as input the structured administrative data, including ICD codes and demographic information from VA-CDW. DeCode addresses feature imputation and label prediction through denoising representation learning. Other data preprocessing steps are described in detail in the supplementary materials.

## Statistical Analysis

DeCode uses the encoder–decoder transformer architecture,[5] which is a better fit to predict future diseases or outcomes than existing deep-learning-based predictive models that use encoder architecture only. We first pretrained DeCode on a large dataset of longitudinal EHRs to predict the complete diagnostic ICD codes of patients' future visits. We add noise such as masking some ICD codes in EHRs to allow DeCode to learn association between ICD codes. We then fine-tuned DeCode for specific single disease or outcome. The four different noising schemes are: code masking, visit permutation, span masking, and visit masking. The specific pretraining objectives and implementation is described in detail in the supplementary materials. We compared DeCode with other strong baseline models: logistic regression, LSTM,[6] and BERT.[7] The prediction of a broad range of disease or outcome onsets are shown in Table 1. Common and rare diseases are selected based on top and least prevalence in VA-CDW.

| Task | Disease or Outcome (ICD-10 code) |
|---|---|
| Prediction of single disease and outcome onset | New onset pancreatic cancer (C25) |
| | Intentional self-harm among patients with PTSD |
| | Avoidable hospital admission for type 2 diabetes or hypertension |
| Disease-agnostic onset prediction for common disease onsets | Chronic post-traumatic stress disorder (F43.12) |
| | Type 2 diabetes (E11.9) |
| | Hyperlipidemia (E78.5) |
| | Loin pain (M54.5) |
| | Low back pain (M54.50) |
| | Obstructive sleep apnea (G47.33) |
| | Depression (F33.9) |
| | Obstructive airway disease (J44.9) |
| | Gastroesophageal reflux disease (K21.9) |
| | Arteriosclerosis (I25.10) |
| Disease-agnostic onset prediction for rare disease onsets | Benign neoplasm of connective tissue of eyelid (D21.0) |
| | Refractory anemia (D46.4) |
| | Melanocytic nevi of upper limb (D22.6) |
| | Benign neoplasm of skin of upper eyelid (D23.10) |
| | Cutaneous abscess of axilla (L02.41) |
| | Ankle and foot subacute osteomyelitis (M86.27) |
| | Cortical hemisphere nontraumatic hemorrhage intracerebral (I61.1) |
| | Malignant neoplasm of head of pancreas (C25.0) |
| | Other complication of kidney transplant (T86.19) |
| | Nonexudative age-related macular degeneration (H35.31) |

**Table 1. Disease or outcome onset** definitions in this study.

## Evaluation Metrics

We used Jaccard similarity score to measure model performance on multi-outcome DAOP. We used positive predictive value (PPV or precision), area under the receiver operating characteristic curve (AUROC), area under the precision recall curve (AUPRC) to measure models' performance on single disease or outcome onset predictions.

**Jaccard similarity score:** Measures the similarity between ground truth and model predictions as the size of the intersection of ground truth and the predictions divided by size of the union. A higher score indicates a better model performance.

**PPV**: PPV is calculated as true positives divided by predicted positives.

**AUROC**: Widely used as an evaluation metric in predictive models, AUROC is the area under sensitivity and false positive rate curve. Sensitivity is the number of true positives divided by the number of ground truth positives. False positive rate is the number of false positives divided by total number of negatives.

**AUPRC**: AUPRC has shown to be an effective measure for highly imbalanced binary classification problems, which include self-harm prediction.[8,9]

## Results

### Disease Agnostic Onset Prediction

DAOP predicts the ICD codes of a patient's future visit based on the ICD codes in the previous visits. In our EHR data, the average number of ICD codes assigned to a patient's visit is 5.18 (st.dev.: 3.79). One baseline model was the state-of-the-art BERT model.[3,10,11] As it is well-documented that physicians copy and paste the content (both structured data and note content) from the previous visits,[12] a second but strong baseline model is to make the prediction by copying the ICD codes of the previous visit.

Table 2 shows that pretraining substantially improves prediction of both common and rare diseases. Compared with the copy baseline model, DeCode improved the Jaccard similarity scores on common diseases by 66.86% (average 30.41 from 45.49 to 75.91) and on rare diseases by 112.96% (average 29.98 from 26.54 to 56.52).

Prediction of rare disease or outcome onsets has been particularly challenging for predictive models. Since DeCode was pretrained on a large EHR dataset, it is able to learn rich representation of clinical variables and their non-linear relations. We therefore speculate that the size of pretraining data may be positively correlated with performance. For verification, we pretrained DeCode on two different EHR sizes: large (6,475,000 patients) and small (64,000 patients). We calculated the relationship between prevalence of ICD codes and performance gain (the accuracy of DeCode pretrained on 6,475,000 patients minus the accuracy pretrained on 64,000 patients). The Pearson's correlation coefficient between performance gain and disease onset prevalence was -0.3161 ($p<0.001$). The ICD codes that showed least performance gain (<10%) included *sleep disorders* (G47), *reaction to severe stress and adjustment disorders* (F43), *type 2 diabetes mellitus* (E11), *disorders of lipoprotein metabolism and other lipidemias* (E78), and *dorsalgia* (M54). The ICD codes with the highest performance gains (>45%) included *personal history of malignant neoplasm* (Z85), *testicular dysfunction* (E29), *other problems related to primary support group including family circumstances* (Z63), *hypertensive heart disease* (I11), *nonrheumatic aortic valve disorders* (I35), and *gout* (M10).

| Models | | Copy | BERT | DeCode Code Masking | DeCode Visit Permutation | DeCode Span Masking | DeCode Visit Masking |
|---|---|---|---|---|---|---|---|
| Chronic PTSD | H | 57.27 | 68.19 | 76.32 | 78.45 | 79.49 | **80.73** |
| | L | 31.21 | 55.55 | 62.71 | 65.81 | 67.03 | **67.52** |
| | 0 | 11.23 | 17.82 | 21.01 | 21.79 | 23.12 | **24.49** |
| Type 2 diabetes | H | 32.42 | 55.97 | 60.98 | 64.57 | 66.45 | **68.57** |
| | L | 28.07 | 51.45 | 53.20 | 57.09 | 58.69 | **59.56** |
| | 0 | 6.52 | 23.39 | 25.03 | 26.87 | 28.76 | **30.50** |
| Hyperlipidemia | H | 31.74 | 57.52 | 60.46 | 66.98 | **68.02** | 67.66 |
| | L | 37.42 | 58.98 | 63.44 | 68.74 | 70.28 | **70.31** |
| | 0 | 8.64 | 22.48 | 24.51 | 30.46 | **32.44** | 30.36 |
| Loin pain | H | 46.81 | 64.77 | 68.75 | 69.59 | 71.52 | **73.03** |
| | L | 39.08 | 56.19 | 64.13 | 65.39 | 67.20 | **68.77** |
| | 0 | 7.33 | 23.90 | 24.40 | 38.53 | 39.84 | **40.07** |
| Low back pain | H | 44.98 | 67.99 | 71.37 | 70.83 | 72.03 | **73.79** |
| | L | 36.22 | 62.61 | 67.20 | 67.70 | 69.58 | **70.62** |
| | 0 | 7.09 | 22.52 | 24.83 | 42.40 | **43.70** | 41.50 |
| Obstructive sleep apnea | H | 58.29 | 76.10 | 84.48 | 84.61 | **86.56** | 86.02 |
| | L | 45.23 | 63.41 | 72.26 | 72.72 | 74.52 | **75.43** |
| | 0 | 4.85 | 23.69 | 31.46 | 40.98 | 37.23 | **42.14** |
| Depression | H | 57.18 | 73.54 | 77.65 | 80.21 | 81.38 | **85.25** |
| | L | 41.95 | 61.55 | 65.85 | 69.61 | 70.62 | **75.19** |
| | 0 | 4.78 | 26.69 | 38.46 | 46.16 | 47.8 | **50.36** |
| Obstructive airway disease | H | 41.70 | 63.83 | 64.59 | 66.65 | 67.70 | **69.74** |
| | L | 30.09 | 51.56 | 56.47 | 59.43 | 61.01 | **64.19** |
| | 0 | 6.07 | 24.88 | 27.38 | 30.82 | **32.79** | 32.22 |
| Gastroesophageal reflux | H | 39.47 | 58.23 | 64.56 | 67.51 | 68.90 | **69.04** |
| | L | 36.41 | 57.17 | 65.10 | 67.69 | 69.31 | **70.45** |
| | 0 | 7.38 | 24.38 | 24.45 | 39.82 | **41.14** | 40.22 |
| Arteriosclerosis | H | 45.05 | 62.17 | 67.63 | 70.77 | 72.73 | **74.64** |
| | L | 37.91 | 59.42 | 62.09 | 64.25 | **65.87** | 65.46 |
| | 0 | 7.12 | 21.81 | 23.25 | 40.26 | 41.48 | **41.72** |
| Rare diseases | 0 | 26.54 | 33.56 | 38.99 | 45.10 | 46.73 | **48.58** |

**Table 2.** Disease-agnostic onset prediction: Jaccard similarity scores on different pretraining objectives for 10 common and 10 rare diseases in Table 1. Many common diseases are chronic in nature. We therefore study whether prior history of the same disease has an impact on prediction performance, where H is high recurrent, L is low recurrent and 0 is new disease onset. We explore various noising schemes (code masking, visit permutation, span masking, and visit masking), and details are included the supplementary materials. "DeCode Visit Masking" is the best performing model. Therefore, the DeCode model refers to the model that incorporated visit masking unless otherwise specified.

Table 2 also shows a comparison of different pretraining objectives for both common and rare diseases. The detail of each objective is described in the supplementary materials. Our results show the importance of autoregressive previous-to-future pretraining, which resulted in a significant improvement (an increase of 4 based on Jaccard similarity) compared to BERT in all prediction subtasks, regardless of common or rare disease or whether the onset was new or recurrent. Our results also show that the visit masking scheme outperformed other masking schemes, resulting in a 3% improvement in Jaccard similarity score on common disease onset predictions and 12% improvement on rare disease onset predictions.

### Single Disease/Outcome Prediction
*Intentional Self-Harm of patients with PTSD*

We also evaluated DeCode for prediction of intentional self-harm among patients with PTSD. This task is challenging because intentional self-harm is a relative rare event (the prevalence was 1.9%) compared to other diseases such as type 2 diabetes onset (the prevalence was 18.7%). As shown in Table 3, AUPRC of DeCode was 16.38 (95% CI, 15.25, 17.52). In contrast, the AUPRC values for BERT, RETAIN, LSTM, and logistic regression models were much lower: 13.17 (95% CI: 11.72, 14.61), 10.54 (95% CI: 9.04, 12.04), 8.32 (95% CI: 7.02, 9.61), 2.81 (95% CI: 0.49, 5.13), respectively, and the improvements from DeCode were all statistically significant. Other metrics (PPV and F1) showed similar results. We also examined how AUPRC changes between different patient demographics. As shown in Supplementary Table 2, AUPRC is stable among different genders, ages, and marital status.

We also examined the impact of how patients' prior history impacts model predictions. We conducted two experiments, called "self-harm with short history" and "self-harm," as shown in Table 3. In the former experiment we only included the 5 most recent visits. In the latter experiment we included all visits (on average, 10) without any reduction of observation window. The AUPRC of DeCode using only the 5 prior visits was 13.82. Using all visits, the AUPRC improved 18% to 16.38, indicating the longer the history, the better the performance.

| Models | Self-Harm w/ Short History | | | Self-Harm w/ Full History | | |
|---|---|---|---|---|---|---|
| | AUPRC | PPV | F1 | AUPRC | PPV | F1 |
| Logistic regression | 6.89 | 3.37 | 6.28 | 2.81 | 2.96 | 5.67 |
| | (1.57) | (0.04) | (0.07) | (1.16) | (0.87) | (1.61) |
| LSTM | 9.14 | 3.92 | 7.30 | 8.32 | 3.71 | 6.88 |
| | (0.56) | (0.25) | (0.43) | (0.65) | (0.13) | (0.29) |
| RETAIN | 9.28 | 5.39 | 8.40 | 10.54 | 5.89 | 8.26 |
| | (0.57) | 0.26 | 0.16 | 0.75 | 1.34 | 1.92 |
| BERT without pretraining | 9.42 | 5.41 | 8.99 | 11.02 | 6.10 | 9.28 |
| | (0.62) | (0.46) | (0.87) | (0.75) | (1.29) | (0.77) |
| BERT | 10.31 | 5.78 | 9.49 | 13.17 | 7.14 | 10.47 |
| | (0.89) | (1.5) | (1.78) | (0.72) | (1.11) | (0.99) |
| DeCode | 13.82 | 6.11 | 10.83 | 16.38 | 8.67 | 14.42 |
| | (0.79) | (0.71) | (0.98) | (0.57) | (0.21) | (0.43) |

**Table 3.** Performance (and standard deviation) of predictive models for intentional self-harm. Each result was calculated from best hyperparameters from 42 trials with 3 randomized seeds each. "Self-Harm w/ Full History" refers to cases where the prediction is based on the original EHR (mean: 10.1 visits, st.dev.: 3.3 visits) prior to predicting intentional self-harm. "Self-Harm w/ Short History" includes only the 5 most recent visits. The first 4 rows are models without pretraining, while the last 2 are models with pretraining.

*Pancreatic Caner*

Results are shown in Supplementary Table 3. DeCode achieved AUROC of 85.35 (95% CI: 83.93-86.77) outperforming the logistic regression model (77.71; 95% CI:75.03, 80.39, p<0.001) and BERT (83.42; 95% CI: 82.98, 83.86, P=0.019).

*Hospital Admission Prediction for Type 2 Diabetes and Hypertension*

For type 2 diabetes, DeCode achieved AUPRC of 50.03 (95% CI: 49.92, 50.13), outperforming the logistic regression model (37.59; 95% CI: 36.51, 38.65, p<0.001) and BERT (49.56; 95% CI: 49.42, 49.70, p<0.001). Compared to BERT without pretraining (BERT initialized randomly and only trained on this task), which achieved AUPRC of 47.15 (95% CI: 46.80, 47.48), BERT with pretraining improved AUPRC by 5.11%, achieving AUPRC of 49.56 (95% CI: 49.42, 49.70, p<0.001). A similar trend was observed for hypertension, as shown in Supplementary Table 4.

## Discussion

In this study, we introduce DeCode, a novel deep neural network model for the prediction of diseases and outcomes using patients' longitudinal EHRs. By first pretraining DeCode on a large EHRs dataset (about 6.5 million patients, a total of 255 million visits from 2016 to 2019) and then fine-tuning on the downstream datasets for specific clinical applications, we show that DeCode outperformed the baseline models on a wide range of both single and multi-disease outcome predictions. As shown in Table 3, the RNN-based model (LSTM), attention-based models (BERT, DeCode), and the hybrid model (RETAIN) all outperformed the logistic regression model. DeCode was the best-performing model and outperformed the logistic regression model on prediction of intentional self-harm (↑ 473% AUPRC, ↑ 183% PPV) and hospital admission for type 2 diabetes and hypertension (↑ 33% AUPRC for type 2 diabetes, ↑ 28% AUPRC for hypertension). The results are not surprising, as deep-learning-based models have improved rich representation of EHR data.[13] In addition, RNN models and attention-based models work well with sequential data such as the natural language.[14]

Our results show the importance of pretraining. For this, we compared the performance between BERT pretrained with a large EHR dataset (about 6.5 million patients) and non-pretrained BERT, whose parameters were randomly initialized before fine-tuned on each task separately. The results, given in Supplementary Table 4, show that pretrained BERT outperformed non-pretrained BERT on hospital admission prediction of both type 2 diabetes (↑ 5.12% AUPRC) and hypertension (↑ 3.03% AUPRC). In addition, pretrained BERT substantially outperformed non-pretrained BERT on intentional self-harm prediction (↑ 19.51% AUPRC) and (↑ 29.55% PPV), as shown in Table 3.

Pretrained models (BERT and DeCode) improve latent representations of EHR data than non-pretrained models. This helps improve probability distribution of candidate diseases or outcomes. As shown in Figure 1 after pretraining on a large EHR cohort, the probability of the next visit ICD code shifted from a random distribution in a model without pretraining to prediction of the correct disease or outcome. While a pretrained model can capture the probability distribution at a large cohort level, fine-tuning can further improve the performance for a domain-specific application. Our results demonstrate the effectiveness of fine-tuning pretrained models.

DeCode outperformed BERT[3,10,11] for prediction of both multiple or single diseases or outcomes, as shown in Tables 2, 3, Supplementary Table 3 and Supplementary Table 4. The performance gain was the highest among rare disease/outcome onset predictions. As shown in Table 2, DeCode improved the Jaccard similarity score by 3% (p<0.001) in predicting onsets of 10 common diseases, and by 16% (p<0.001) in predicting onsets of 10 rare diseases compared with the BERT model with the bidirectional objectives. DeCode also substantially improved AUPRC for intentional self-harm prediction (from 13.17 to 16.38, p<0.001), AUROC for pancreatic cancer prediction (from 83.42 to 85.35, p<0.001), and AUPRC for hospital admission prediction for type 2 diabetes (from 49.56 to 50.03, p<0.001). On the other hand, the improvements of AUROC for PTSD (from 80.63 to 80.91, p=0.5808) and AUPRC for hypertension admission (from 37.78 to 36.72, p=0.1482) were minimal. Our results also show that the performance gains by DeCode were negatively correlated with the prevalence of disease/outcome, indicating that DeCode is relatively better at predicting rare disease/outcome onset.

Attention-based models benefit from longitudinal EHRs with long histories. As shown Table 3, of the attention-based models (DeCode and BERT) trained on longer histories, the AUPRC scores improved 18% and 27%, respectively, in comparison with models trained on only the 5 most recent visits. In contrast, the AUPRC scores of non-attention-based models such as LSTM and logistic regression decreased 8% and 58%, respectively, when trained on longer histories (about 10 prior visits) compared with the models trained on only the 5 most recent visits. These findings are consistent with the previous AI research,[15] which shows that LSTM, although mitigating the vanishing gradient descent challenge of RNN, remains suboptimal with long-time dependencies.

Our work is also related to predictive model studies focused on intentional self-harm.[16–20] Typically, these approaches sample thousands of patient and use probability tables, decision trees, and logistic regression to predict intentional self-harm. Hartl et al.[18] is the only study that predicts intentional self-harm among PTSD patients (prevalence ratio 5%, PPV 0). Simon et al.[20] integrated EHR data and questionnaires on 2,960,929 patient to predict suicide attempts (prevalence ratio 1%) within 90 days of a mental health visit. The most successful model demonstrated a PPV of 5%. In comparison, we collected a large cohort of 70,623 PTSD patients. We also did not use any downsampling technique to balance the positive and negative data.

DeCode outperformed other models to achieve a PPV of 8.67% for prediction of intentional self-harm among PTSD patients (prevalence ratio 1.9%, details in the Results section). We also calculated operating characteristics, including specificity, sensitivity, and PPV for a variety of thresholds. As shown in Supplementary Table 5, DeCode has a sensitivity of 42.25% for patients in the top-10% of predicted risk. This suggests that if an intensive case management program were provided with the 10% of PTSD patients with highest predicted self-harm risk, it would include 42.25% of the patients who would self-harm for the first time between this visit and next visit. In comparison, the sensitivity at the 10% threshold for logistic regression is only 10.95%. The proportion of self-harm noncases that were below the top-10% of predicted risk (specificity) of DeCode is 99.16%. In comparison, the specificity at the 10% threshold for logistic regression is only 90.75%. The percentage of patients receiving the intervention who would otherwise attempt self-harm for the first time (PPV) is a crucial metric in estimating the potential benefit of any intensive case management intervention. The highest PPV for DeCode is 8.45% for the 10% threshold. In other words, out of the 100 highest-predicted-risk patients from 1,000 previously diagnosed PTSD patients, 8.45 patients would attempt intentional self-harm for the first time before the next visit. In comparison, for logistic regression the PPV at the 10% threshold is 2.17%. A practical suicide prevention tool must have a relatively high PPV so as minimize the resources and intrusion directed at patients who will never attempt self-harm.[21]

Pretrained models have shown to be able to learn commonsense knowledge in the open domain. We therefore hypothesize that DeCode is able to learn commonsense medical knowledge through pretraining. We adapted the bertviz tool to visualize how attentions work in DeCode. We randomly selected a patient who was diagnosed with PTSD and suffered from multiple visit of low back pain and visualized the attention from all 12 layers and 12 heads learned by DeCode. Each visit is separated by "[SEP]" token. As shown in subfigure b. of Figure 2, when predicting whether patient is diagnosed as PTSD in the last visit, attention head5 at decoder layer3 (purple) attends both unspecified low back pain (ICD-10: M54.50) when predicting PTSD (ICD-10: F43.12). However, when only one unspecified low back pain is presented in history, then this attention to PTSD disappears as shown in the subfigure a. In literature, multiple studies have shown a high correlation between chronic pain and PTSD. Compared to 4-12% PTSD in the general public, more than 30% of pain patients are suffering from comorbid PTSD. And low back pain is a common physical symptom in comorbid PTSD. Figure 2 shows that DeCode attends low back pain to PTSD more from later visit than that from the previous visit, indicating the effect of multiple visits of low back pain instead of just once low back pain. In other words, this potentially indicates that DeCode learns the longitudinal relationship between visits. We also verified this attention with other similar patients and found that head5 always attends low back pain to PTSD, which shows that this attention is independent of contextual noise.

## Future Works

There are multiple future research directions. First, DeCode is currently trained only with diagnostic ICD codes. Other information such as procedure codes, medications, lab results, and phenotypical information extracted from notes can be added to further improve performance. Second, our prediction of intentional self-harm focuses on patients with PTSD. Future work we will also evaluate predictions among other subgroups for prediction of suicide death.

## Conclusions

In conclusion, we proposed DeCode, a bidirectional and autoregressive model pretrained on a large volume of structured EHRs, and further evaluated the model in various disease or outcome onsets predictions. The results of this prognostic study suggest that DeCode has higher prediction accuracy compared to previous machine learning models on rare disease or outcome onsets, such as predicting intentional self-harm of patients with PTSD.

## Data Availability

The EHR data cannot be shared publicly due to HIPAA regulation. The machine learning algorithms and software will be made available upon acceptance.

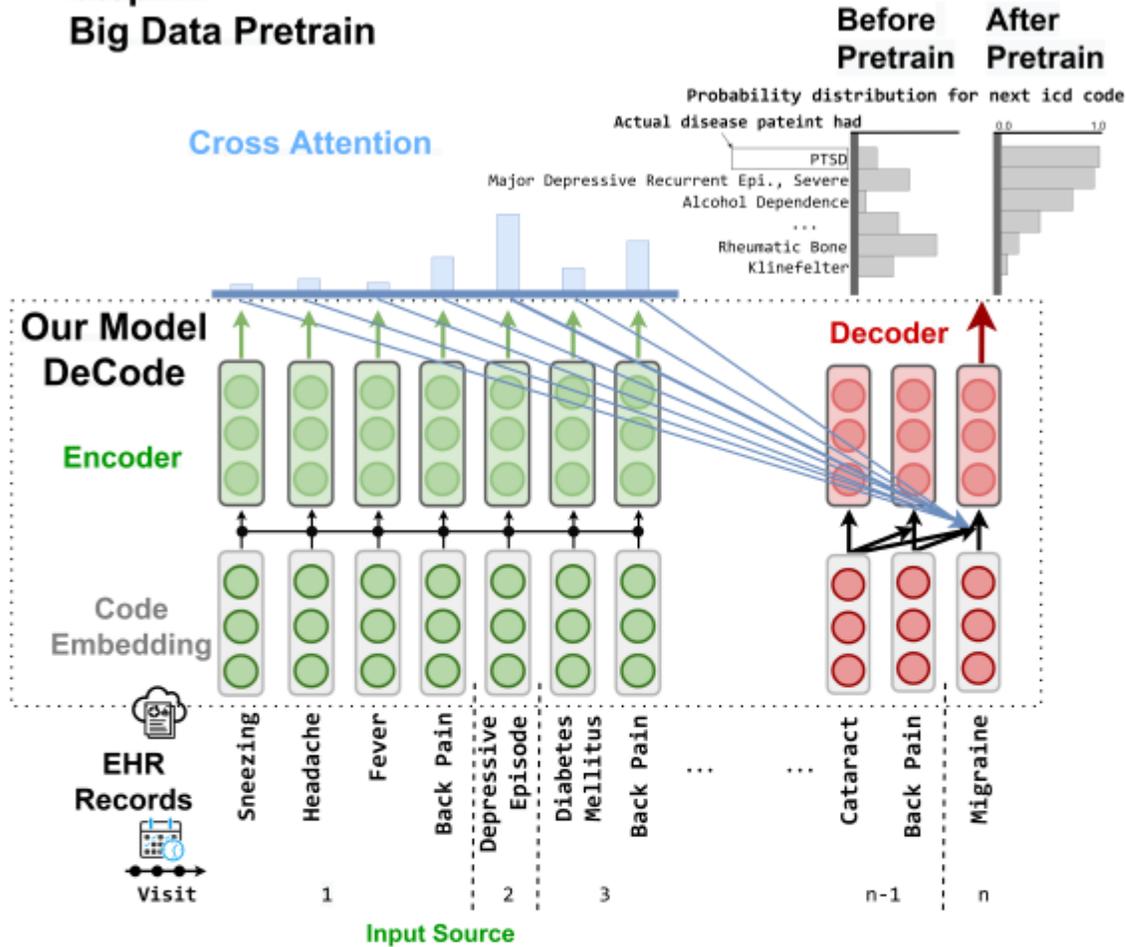

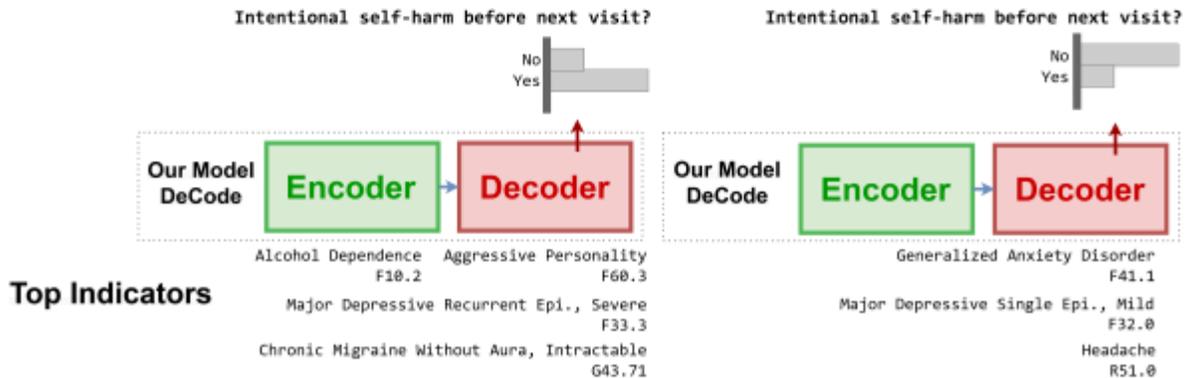

**Figure 1.** DeCode architecture and pretrain/fine-tune pipeline. DeCode was first pretrained with bidirectional encoder and autoregressive previous-to-future decoder on large longitudinal EHR data. In pretraining, DeCode learns the probability distribution of ICD codes (vs random distribution) at visit level. We then fine-tuned DeCode to the downstream predictions of single diseases or outcomes. Through attention weights, DeCode was able to identify risk factors for the predictions. An interactive figure could be found here: https://icd-demo.herokuapp.com. Note that in this study, although we only pretrained DeCode on the ICD codes, DeCode could be trained on any EHR data types.

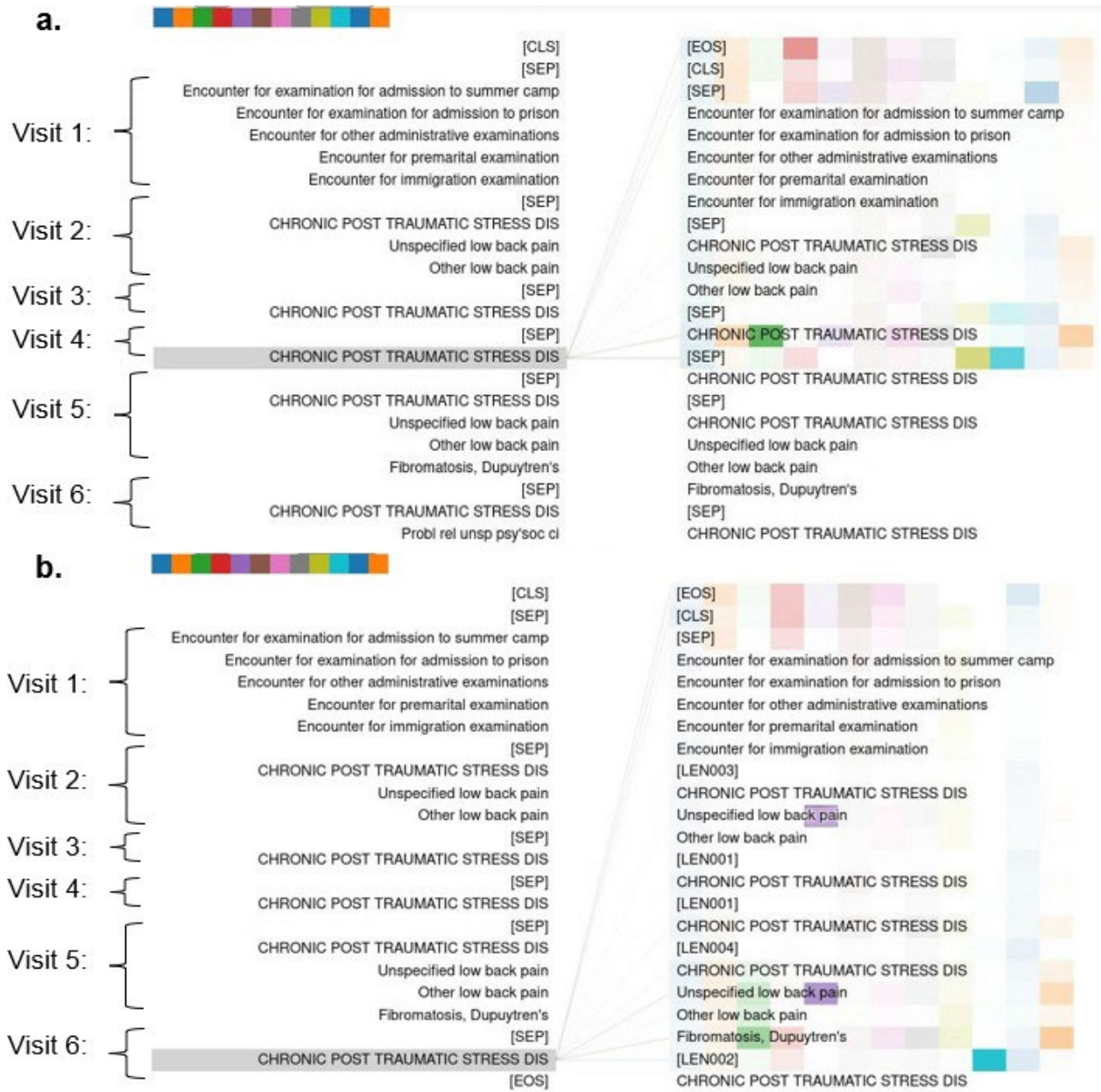

**Figure 2.** Disease agnostic onset prediction: An example of how decoder layer3 head5 attention in **purple** attends between PTSD and multiple visits of low back pain (ICD-10: M54.50) but not between PTSD and one (non-chronic) low back pain. **a**: predicting PTSD when patient has only one previous low back pain. **b**: predicting PTSD when patient has low back pain from multiple previous visits. Color of the block represents different heads, and intensity of the block represents attention weights of the specific head. Comparing **b** to **a**, **purple** box disappears, while other boxes stays relatively the same.